\documentclass[runningheads]{llncs}

\usepackage[a4paper, margin=1in]{geometry}
\usepackage[]{eccv}

\usepackage{eccvabbrv}

\usepackage{graphicx}
\usepackage{booktabs}
\usepackage{multirow}
\usepackage[accsupp]{axessibility}  

\usepackage[pagebackref,breaklinks,colorlinks,citecolor=eccvblue]{hyperref}

\usepackage{orcidlink}

\begin{document}

\title{15M Multimodal Facial Image-Text Dataset} 


\author{Dawei Dai \and
YuTang Li \and
YingGe Liu \and
Mingming Jia \and
Zhang YuanHui \and
Guoyin Wang
}

\authorrunning{Dawei Dai, Yutang Li et al.}

\institute{College of Computer Science and Technology, Chongqing University of Posts and Telecommunications, China\\}

\maketitle
\begin{figure}[h]
	\centering
	\includegraphics[width=14cm]{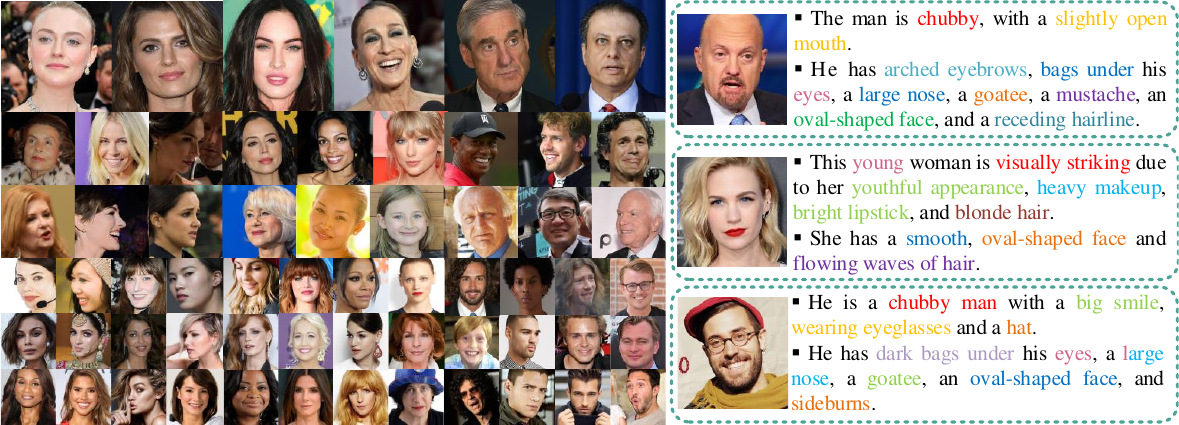}
	\caption{Overview of our proposed FaceCaption-15M containing over 15 million facial image-text (right and left) pairs.}
	\label{fig1}
\end{figure}  

\begin{abstract}
  Currently, image-text-driven multi-modal deep learning models have demonstrated their outstanding potential in many fields. In practice, tasks centered around facial images have broad application prospects. This paper presents \textbf{FaceCaption-15M}, a large-scale, diverse, and high-quality dataset of facial images accompanied by their natural language descriptions (facial image-to-text). This dataset aims to facilitate a study on face-centered tasks. FaceCaption-15M comprises over 15 million pairs of facial images and their corresponding natural language descriptions of facial features, making it the largest facial image-caption dataset to date. We conducted a comprehensive analysis of image quality, text naturalness, text complexity, and text-image relevance to demonstrate the superiority of FaceCaption-15M. To validate the effectiveness of FaceCaption-15M, we first trained a facial language-image pre-training model (FLIP, similar to CLIP) to align facial image with its corresponding captions in feature space. Subsequently, using both image and text encoders and fine-tuning only the linear layer, our FLIP-based models achieved state-of-the-art results on two challenging face-centered tasks. The purpose is to promote research in the field of face-related tasks through the availability of the proposed FaceCaption-15M dataset. All data, codes, and models are publicly available. \href{https://huggingface.co/datasets/OpenFace-CQUPT/FaceCaption-15M}{https://huggingface.co/datasets/OpenFace-CQUPT/FaceCaption-15M}
\end{abstract}

\section{Introduction}
\label{sec:intro}

In recent years, large-scale data-driven pre-training models have achieved significant success in fields of computer vision and natural language processing \cite{du2022glm,taori2023alpaca,NEURIPS2022_b1efde53, Antol_2015_ICCV, Rombach_2022_CVPR, Esser_2021_CVPR, Zhang_2023_ICCV}. In practical applications, tasks centered around faces have broad application across various aspects, such as social media, mobile payment systems and security monitoring. These applications heavily rely on the availability of datasets, such as ImageNet \cite{imagenet_cvpr09}, CelebA \cite{liu2015faceattributes}, CelebA-HQ \cite{7d745f7015ea47f89e07bd963e573213}, LAION-Face \cite{zheng2022general}, FFHQ \cite{Karras_2019_CVPR} and CelebV-Text \cite{yu2023celebv}.

A multimodal image-text model exhibit the potential to achieve better generalization capabilities. However, training a multi-modal model for aligning face image-text remains challenging owing to the limitations in existing open-source face image datasets. These limitations include the weak correlation between text and facial images, and a shortage of samples. A large-scale, high-quality facial image-text dataset can significantly promote the efficient development of face-related research. In this work, “large-scale” refers to the dataset size that meets the requirements for training large models, such as CLIP, with tens of millions or more samples; “high-quality” refers to a dataset primarily comprising facial images with minimal background interference; “face captions” refer to natural language descriptions of human facial features.

Constructing a large-scale and high-quality facial image-text dataset presents several challenges \cite{zheng2022general,7780896,Karras_2019_CVPR}, mainly in three aspects: (1) Obtaining a high-quality facial image dataset comprising tens of millions of images while ensuring a natural distribution and precise alignment of face. (2) Ensuring the relevance of text to facial images requires the use of natural language to describe intricate facial details. (3) Text generation: Manual generation is both expensive and non-scalable, while automatic methods, although scalable, often suffer from problems such as semantic bias and high repetition.

To address the above challenges, we developed a comprehensive process to construct our FaceCaption-15M dataset. An overview of our approach: First, we followed the steps similar to those used in CelebV-HQ \cite{zhu2022celebvhq} to obtain the images containing faces from LAION-Face \cite{zheng2022general}, and then RetinaFace \cite{9157330} was used to detect and crop the regions of facial images. Second, to ensure a high correlation between image and text, we used the method proposed by He et al. \cite{10.1145/3123266.3123424} to predict corresponding facial attributes. Third, we designed an automatic method combining grammatical templates and large-scale language model (LLM) to generate the caption for facial image.

\begin{table*}[!t]
	\centering
	\scriptsize
	\caption{Comparisons with other popular facial image datasets. Symbol “\#” indicates the number of samples (images or image-text pairs). Abbreviations “mRes”, “Ann”, and “mWords” denote average resolution of all images, the number of annotations, and average words of all text, respectively. Abbreviation “Align” indicates whether the image only contains faces.}
	\setlength{\tabcolsep}{.6mm}{
		\begin{tabular}{@{}c|cccccccc@{}}
			\toprule[1pt]
			\multirow{2}{*}{Datasets}  & \multicolumn{4}{c}{Image/Video}                                                          & \multicolumn{4}{c}{Caption/Text}                     \\ \cline{2-9} 
			& \#Samples    & mRes.              & \#Ann.       & \multicolumn{1}{c|}{Align}      & \#Samples    & \#mWords    & Natural    & Relevance  \\ \toprule[1pt]
			LFWA \cite{5674057}                 & 13K          & 250*250            & 40           & \multicolumn{1}{c|}{\checkmark}          & -            & -           & -          & -          \\
			CelebA \cite{liu2015faceattributes}               & 202K         & 178*218            & 40           & \multicolumn{1}{c|}{\checkmark}          & -            & -           & -          & -          \\
			FFHQ-Text \cite{zhou2021generative,karras2019style}            & 0.76K        & \textbf{1024*1024} & \textbf{162} & \multicolumn{1}{c|}{\checkmark}          & 6.8K         & 22          & \checkmark          & \checkmark          \\
			MM-CelebA \cite{Xia_2021_CVPR}           & 30K          & 512*512            & 38           & \multicolumn{1}{c|}{\checkmark}          & 300K         & 17          & \checkmark          & \checkmark          \\
			CelebA-Dialog \cite{jiang2021talkedit}        & 202K         & 256*256            & 5            & \multicolumn{1}{c|}{\checkmark}          & 202K         & 25          & \checkmark          & \checkmark          \\
			CelebV-Text  \cite{yu2023celebv}        & 70K         & 512*512$+$            & 77            & \multicolumn{1}{c|}{\checkmark}          & 1.4M         & 67          & \checkmark          & \checkmark          \\
			LAION-Face \cite{zheng2022general}           & \textbf{50M} & 615*615            & -            & \multicolumn{1}{c|}{$\times$}          & \textbf{50M} & 12          & $\times$      & $\times$          \\
			\textbf{FaceCaption-15M} & \textbf{15M} & \textbf{285*285}   & \textbf{40}  & \multicolumn{1}{c|}{\textbf{\checkmark}} & \textbf{15M} & \textbf{30} & \textbf{\checkmark} & \textbf{\checkmark} \\ \toprule[1pt]
		\end{tabular}%
	}	
	\label{table1}
\end{table*}

Our FaceCaption-15M contains over 15 million aligned facial image-text pairs. Some illustrations are presented in \Cref{fig1}. \Cref{table1} shows a comparison between FaceCaption-15M and other renowned facial image datasets, demonstrating its superiority in terms of quantity, the naturalness of text, and the relevance of image-text. To further validate its superiority, we first trained a CLIP-like model (named FLIP) on the FaceCaption-15M as a pre-trained model. Subsequently, we applied the FLIP model to three facial image-related tasks, including facial attribute recognition, text-image retrieval and sketch-based facial image retrieval. \textbf{Our contributions can be summarized as follows:}

(1) We propose FaceCaption-15M, the largest facial image-text dataset featuring high-quality images and rich, highly relevant caption. (2) We conduct a comprehensive statistical analysis to assess the quality and diversity of face images and text, and the correlation between text and face images, all of which demonstrate the superiority of our FaceCaption-15M. (3) We train a facial language-image pre-trained model on the FaceCaption-15M dataset and apply it to two face-related tasks, achieving state-of-the-art results.

\section{Related Work}
\subsection{Face-related Tasks}
Facial images have given rise to numerous practical applications. For example, image retrieval \cite{ZHANG2023109671,9320210,10095094}, face recognition \cite {Kim_2022_CVPR,10.1007/978-3-031-19775-8_41}, facial image generation \cite{Deng_2019_CVPR,Richardson_2021_CVPR, Shen_2020_CVPR}, facial image parsing \cite{zheng2022general, Zheng_2022_CVPR, LIN2021104190} and editing \cite{Huang_2023_CVPR, Shen_2020_CVPR, Jo_2019_ICCV, Sun_2022_CVPR}. For task of facial attribute recognition, many studies \cite{Deng_2019_CVPR, boutros2022elasticface, mare2021realistic, schroff2015facenet} have been explored to improve the accuracy, which are primarily based on datasets such as CelebA \cite{liu2015faceattributes} and LFWA \cite{5674057}, with widespread applications across multiple fields. For sketch-based facial image retrieval task, widely used datasets include FS2K \cite{Fan2022FS2K} and CUFSF \cite{zhang2011coupled}. Efforts \cite{10095094, chen2019semi,fan2019scoot, Fan2022FS2K, 10.5555/3618408.3619075} in this area mainly focus on (1) learning efficient representations for sketch and image. This task holds great potential applications, including digital entertainment and suspect identification. In recent years, the way of fine-tuning a pretrained models has demonstrated significant advantages \cite{Baldrati_2022_CVPR,9956714,delmas2023posefix,shen2023clip,lou2022tecm}.

\subsection{Datasets}
\textbf{Single-modal datasets.} Single-modal face datasets mainly consist of images, which played a pivotal role in early research. For example, CelebA \cite{liu2015faceattributes} and LFWA \cite{5674057} datasets provide a large number of facial attribute annotations; VGGFace2 \cite{cao2018vggface2} and CASIA-WebFace \cite{yi2014learning} datasets were specifically designed for face recognition tasks. They include a large number of celebrity face images collected from the Web, as well as face images captured under different conditions, such as different poses, expressions, and lighting conditions.

\noindent\textbf{Multi-modal datasets.} 
The Flickr30k \cite{young2014image} dataset includes approximately 31,000 facial images collected from Flickr, each of which has been annotated using five reference sentences by human annotators. However, these images often feature complex backgrounds, and the corresponding text do not naturally capture facial features. Although MM-CelebA \cite{Xia_2021_CVPR} and CelebA-Dialog \cite{jiang2021talkedit} contain multiple pairs of face description labeled by humans, the sample sizes are insufficient, making them inadequate for training a large model. LAION-Face \cite{zheng2022general} dataset, a subset of LAION-400M \cite{DBLP:journals/corr/abs-2111-02114}, stands as the largest existing face image-text dataset, containing approximately 50M pairs of image-text. However, the text within this dataset was directly extracted from the Internet and exhibit very weak correlation with the face. Moreover, the images often contain complex backgrounds, which may not suffice to support delicate facial-related tasks.

\noindent\textbf{In Summary.} Owing to the lack of large-scale and high quality facial image-text datasets, researchers often first train a model (such as ResNet \cite{He_2016_CVPR}, VIT \cite{dosovitskiy2020vit} and CLIP \cite{pmlr-v139-radford21a}) on the general large-scale datasets such as LAION-5B \cite{NEURIPS2022_a1859deb}, CC \cite{sharma2018conceptual}, ImageNet22K \cite{ILSVRC15}, and COCO \cite{lin2014microsoft} as pre-trained modules. Subsequently, they fine-tune the pre-trained models on a smaller-scale dataset for specific facial-related tasks. However, these pre-trained models tend to exhibit insufficient generalization capabilities on facial-related tasks. Overall, various limitations underscore the pressing need for a large-scale, high-quality, multi-modal facial dataset featuring natural language descriptions of facial details to support more complex facial-related tasks \cite{Wasim_2023_CVPR,pmlr-v162-li22n}.

\section{Dataset Construction}
\label{3}
\begin{figure*}[!t]
	\centering
	\includegraphics[width=.98\linewidth]{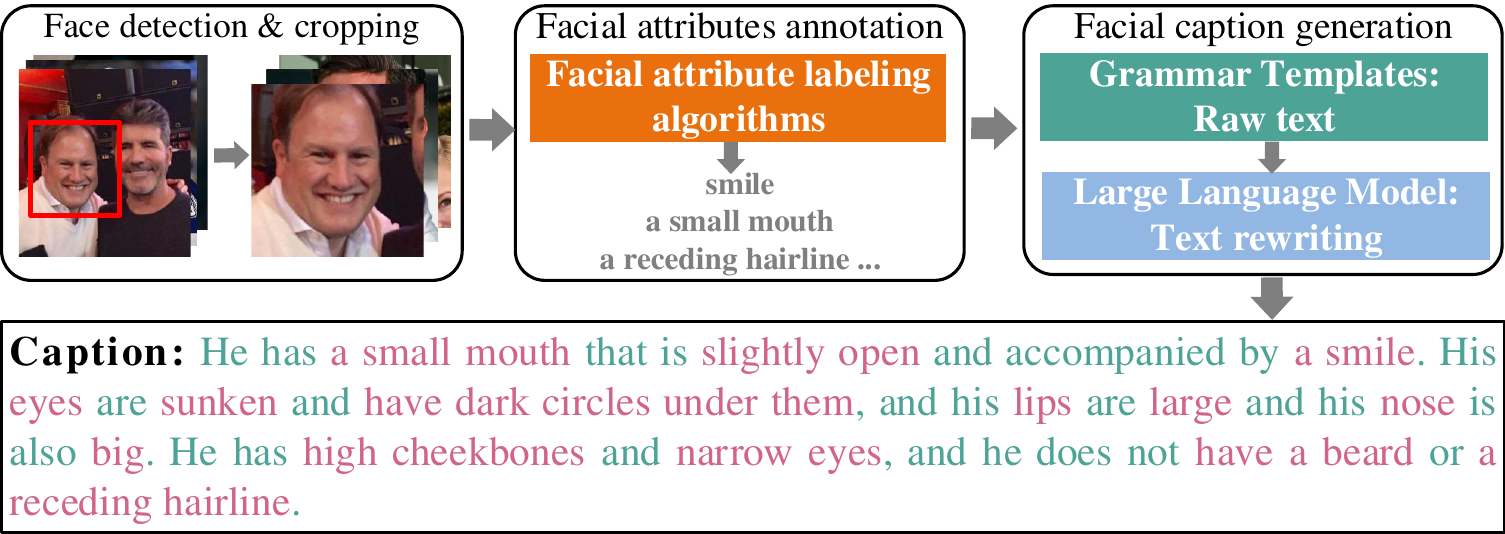}	
	\caption{Pipeline of our FaceCaption-15M construction process. Pipeline includes face detection and cropping, face annotation, and automatic caption generation.}
	\label{fig2}
\end{figure*}

Our purpose is to construct a large-scale and high-quality facial image-text dataset featuring natural language descriptions that accurately describe facial features. The general approach of building such a large-scale facial image-text dataset is to first collect as many images containing people as possible, then select an excellent-performing algorithm to segment the face part of the image, and finally select or design a suitable algorithm to generate high-quality text descriptions for the face images \cite{Karras_2019_CVPR,yu2023celebv,jiang2022text2human}. As shown in \Cref{fig2}, for the construction of our FaceCaption-15M.

\subsection{Facial Images Collection}
\textbf{Image Collection.} Specifically, we accessed the LAION-Face \cite{zheng2022general} dataset, which contains over 50M image-text pairs that obtained through web crawling, as our source of raw data. LAION-Face is of a considerable scale, and its image distribution closely resembles real-world distributions. Moreover, using such a dataset as our raw data source offers significant cost savings compared to manual data collection. There were limitations stemming from link expiration and network issues, as we could only access about 75\% of the LAION-Face. 

\noindent\textbf{Face Segmentation.} For original LAION-Face dataset, we segment the image of the facial regions. First, we selected all images with faces from LAION-Face using RetinaFace \cite{9157330} model, which resulted in approximately 37M images. To obtain a high-quality facial image dataset while avoiding noise interference, we conducted cropping, alignment, and filtering of the facial images based on facial region detection boxes. Specifically, we retained only those facial regions with resolutions exceeding 128 $\times$ 128 pixels and confidence scores higher than 0.98, resulting in approximately 23M images. Importantly, to maintain image quality, we did not uniformly scale the images to the same size, resulting in varying resolutions among the collected images.

\subsection{Facial Attributes Annotation}
\label{3.2}
Attributes play a pivotal role in generating the description text for facial image, thereby determining the correlation between the image and text. We designed 40 appearance attributes for facial features. Given the considerations of annotating a vast amount of data, we selected an automatic annotation method. In terms of efficiency and accuracy, we employed an open-source algorithm \cite{10.1145/3123266.3123424} for predicting image attributes. To enhance the reliability of annotations, we retained only the labels predicted by the model with a probability exceeding 0.85. Furthermore, to generate more accurate natural language text, we retained samples with more than five valid predicted labels. Finally, we reduced the dataset size to 15M.

\subsection{Facial Caption Generation: Raw Text Generation and Rewriting}
Since, image-text pairs in LAION-Face dataset were obtained through subtitle crawling, and the text showed a weak correlation with the accompanying image. Our aim is to generate the caption of facial images. The manual annotation, while accurate, is time-consuming and labor-intensive, making it unsuitable for constructing large-scale datasets. However, automatic methods often offer efficiency and scalability. Nevertheless, the diversity, complexity, and naturalness of description sentences generated by traditional automatic text generation methods are limited by grammatical templates. With the development of LLM \cite{NEURIPS2022_b1efde53,10.1155/2022/7839840,du2022glm,qwen,baichuan2023baichuan2,touvron2023llama}, text generated by these models is endowed with high diversity and naturalness. Here, we propose a text generation strategy that combines grammatical templates with LLM. Specifically, (1) we first input the attribute annotations generated by \Cref{3.2} into the designed grammatical template to generate the raw text, and then (2) we input the raw text into the LLM to generate natural, diverse, and accurate text descriptions.

To ensure the generation of high-quality description text using LLM, the quality of the raw text generated by the grammatical template is paramount. Here, we adopted the probabilistic context-free grammars (PCFG \cite{Suppes1972}) algorithm to generate the raw text as multiple short sentences, each constructed using different attributes. The performance of the LLM model itself affects the quality of the generated caption. We conducted research on existing open-source LLMs and finally selected the Qwen-7B-Chat model \cite{qwen}.

\section{Statistical Analysis for FaceCaption-15M Dataset}
In this section, we conducted a comprehensive analysis for our FaceCaption-15M dataset. To provide an overview, we compared the overall statistics of FaceCaption-15M with some existing popular face datasets, as presented in \Cref{table1}. More comparisons are as the following sections.

\subsection{Image Quality Comparisons}
We first conducted a quantitative comparison of image quality across some facial image datasets: CelebA-Dialog \cite{jiang2021talkedit}, MM-CelebA \cite{Xia_2021_CVPR}, CelebV-Text \cite{yu2023celebv} (randomly selecting 10 frames from each video to evaluate their quality), and LAION-Face \cite{zheng2022general}. We employed two general no-reference image quality assessment methods, BRISQUE \cite{Mittal_Moorthy_Bovik_2012} and CLIPIQA \cite{wang2022exploring}, to evaluate the image quality within these datasets. The BRISQUE method evaluates image quality by calculating the local normalized brightness coefficient of the image pixels, where lower scores indicate better image quality. CLIPIQA method \cite{DBLP:conf/icml/RadfordKHRGASAM21} assesses image quality by calculating the cosine similarity between the image and predefined prompts, with higher scores indicating better image quality.

\begin{figure*}[!t]
	\centering
	\includegraphics[width=1\linewidth]{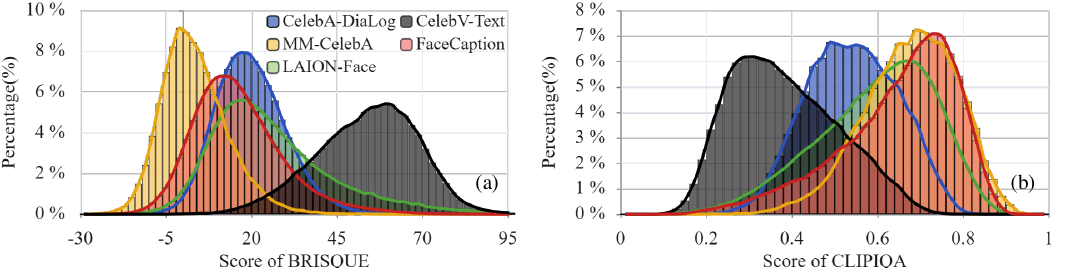}	
	\caption{Image quality score distribution. (a) BRISQUE \cite{Mittal_Moorthy_Bovik_2012} evaluation with lower scores indicating better image quality; (b) CLIPIQA \cite{wang2022exploring} evaluation with higher scores indicating better image quality.}
	\label{fig3}
\end{figure*}

Specifically, we assigned scores to each image in the five datasets and calculated the proportions of different scores within each dataset. As shown in \Cref{fig3}(a), it is evident that the image quality score distribution of our FaceCaption-15M is better than that of LAION-Face, CelebV-Text and CelebA-Dialog, but falls slightly behind MM-CelebA, as assessed by the BRISQUE \cite{Mittal_Moorthy_Bovik_2012} evaluation. This is because some images in MM-CelebA are sourced from the higher-quality face dataset CelebAMask-HQ \cite{DBLP:conf/cvpr/Lee0W020}. As shown in \Cref{fig3}(b), the image quality score distribution of FaceCaption-15M is observed to be close to that of MM-CelebA and notably better than that of LAION-Face, CelebV-Text and CelebA-Dialog, considering the CLIPIQA \cite{wang2022exploring} evaluation.

\begin{figure}[htpb]
	\centering
	\includegraphics[width=1\linewidth]{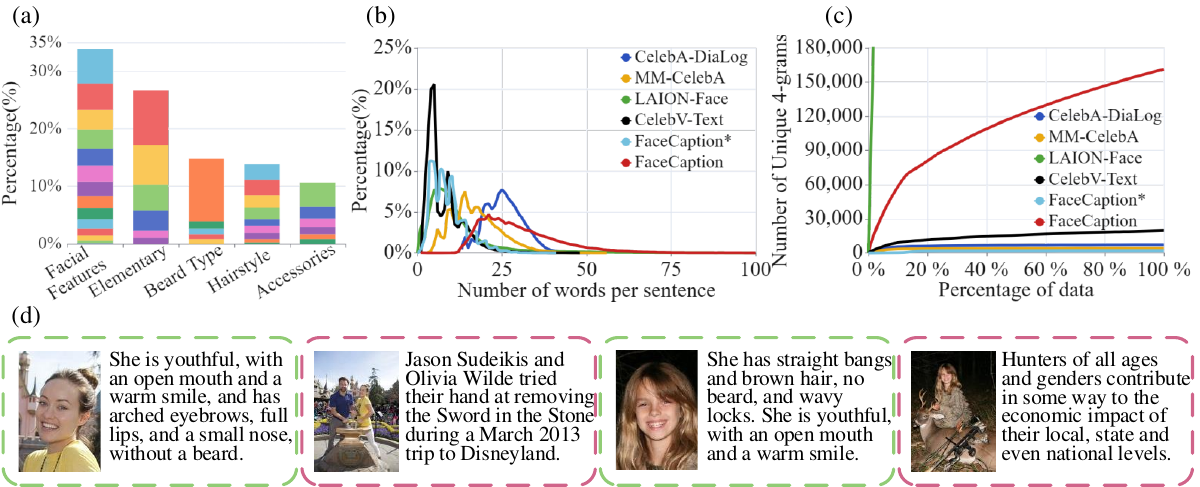}	
	\caption{Text distribution. (a) Distribution of the five categories of annotations in the FaceCaption-15M. (b) The percentage of sentences in the dataset with different word counts. (c) The number of unique 4-grams under the percentage data. (d) Illustrations of image-text pairs LAION-Face and FaceCapition-15M. \textbf{FaceCaption* indicates the caption generated by grammatical template without using LLM}.}
	\label{fig4}
\end{figure}

\subsection{Text Comparison}
Firstly, we performed a statistical analysis of the attribute annotations within FaceCapition-15M. As shown in \Cref{fig4}(a), we divided the 40 facial-appearance attributes into five categories. Among these categories, facial characteristics (such as a double chin and big nose) account for approximately 33.9\%. The elementary characteristics of the portrait (including young, male, and blurry) accounted for approximately 26.7\%, while beard and hair characteristics each account for approximately 14.9\% and 13.9\%, respectively. Accessory characteristics made up 10.6\% of the annotations.

In comparison to CelebA-Dialog \cite{jiang2021talkedit}, MM-CelebA \cite{Xia_2021_CVPR}, CelebV-Text \cite{yu2023celebv} and LAION-Face \cite{zheng2022general}, the text within FaceCaption-15M are more extensive and detailed (See \Cref {fig4}(b)). Specifically, the average text lengths for CelebA-Dialog, MM-CelebA, LAION-Face, and FaceCaption-15M are 25, 17, 12, and 30, respectively. As shown in \Cref{fig4}(c), to further evaluate the naturalness and complexity of the text in FaceCaption-15M, we calculated the type-token vocabulary curve \cite{Youmans_1990} for all texts. Here, unique 4-grams represent the combination of all unique four consecutive words in the corpus, with a larger value reflecting the higher naturalness and complexity of the language \cite{Wang_Wu_Chen_Li_Wang_Wang_2019}. Owing to the incorporation of our grammar templates and the subsequent rewriting by LLM (FaceCapition* $vs$ FaceCapition), the naturalness and complexity of FaceCaption-15M surpassed those of MM-CelebA, CelebA-Dialog and CelebV-Text significantly. It is worth noting that LAION-Face exhibited even greater naturalness and complexity than FaceCaption-15M. This discrepancy arises because the text in LAION-Face is directly sourced from the Internet and is not constrained by any specific format \cite{zheng2022general}. As shown in \Cref{fig4}(d), both image and text within FaceCaption-15M dataset focuses more on the correlation between text and facial details comparing with the LAION-Face.

\begin{figure}[!t]
	\centering
	\includegraphics[width=.65\linewidth]{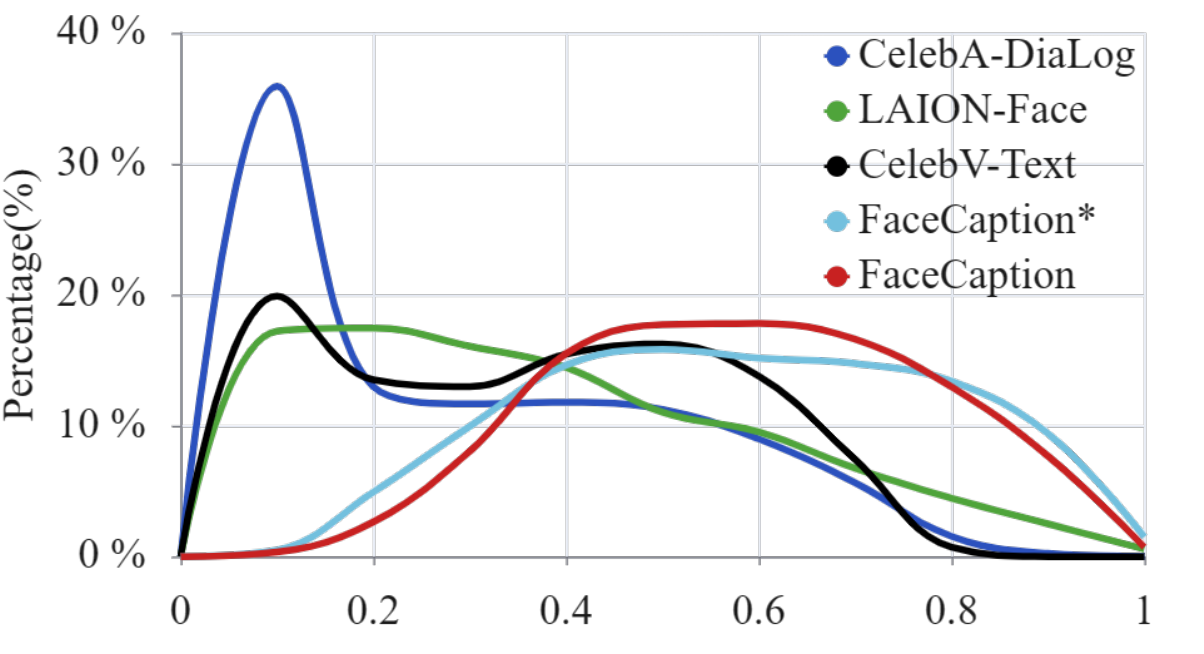}	
	\caption{Image-text matching score. We adopt the ITM score to measure image-text correlation of different datasets.}
	\label{fig6}
\end{figure}

\subsection{Text-Image Relevance Comparison}
To quantitatively verify the correlation between facial images and text, we adopted the image-text matching (ITM) score \cite{DBLP:conf/nips/LiSGJXH21,pmlr-v162-li22n} as a metric to measure the degree of alignment between the two modalities. A larger ITM score indicates a greater correlation between text and images. In this regard, we selected an equal number of samples from the CelebA-Dialog \cite{jiang2021talkedit}, LAION-Face \cite{zheng2022general}, CelebV-Text \cite{yu2023celebv} (randomly selecting some frames from each video), FaceCaption* and FaceCaption-15M datasets to fine-tune the last linear layer of BLIP \cite{pmlr-v162-li22n}, obtaining five new BLIP models. These models were then used for the text-based image retrieval task on the test set of MM-CelebA \cite{Xia_2021_CVPR}, which exhibits a high correlation between text and images within FaceCapition* and FaceCapition. As shown in \Cref{fig6}, the distribution of ITM scores in FaceCaption-15M surpassed that of CelebA-Dialog, CelebV-Text and LAION-Face. Notably, the distribution of ITM scores in CelebA-Dialog performed the worst, further proving that FaceCaption-15M exhibits a high correlation between the descriptions and facial images.

\begin{figure}[htpb]
	\centering
	\includegraphics[width=0.96\linewidth]{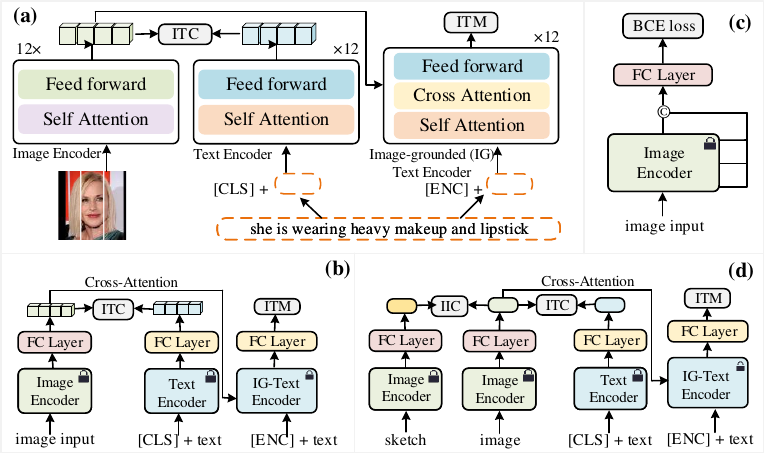}	
	\caption{Overview of FLIP architecture (a). Same color represents shared parameters. “12x” stands for 12-layer transformer modules. (b), (c) and (d) FLIP-based model are applied to the tasks of text-image retrieval, facial attributes prediction and sketch less facial image retrieval, respectively.}
	\label{fig7}
\end{figure}

\section{Experiment}
\subsection{FLIP: Facial Language-Image Pre-training Model}
\noindent\textbf{Architecture:} \noindent Based on FaceCaption-15M, we trained a multimodal representation model (FLIP), similar in concept to CLIP \cite{DBLP:conf/icml/RadfordKHRGASAM21}, designed for aligning facial images with semantics. As shown in \Cref{fig7}(a), FLIP contains the following components: (1) Image Encoder \cite{dosovitskiy2020vit}: Composed of a visual transformer, this component processes the image. (2) Text Encoder: When handling text input alone, this encoder follows the standard BERT \cite{DBLP:conf/naacl/DevlinCLT19} module and uses the [CLS] token to summarize the entire sentence. In the case of multimodal input, a cross-attention layer is introduced between the self-attention layer and the feed-forward network of the text encoder to fuse visual information (Image-grounded Text Encoder). To adapt to specific tasks, an [ENC] token is added to the text input, serving as the multimodal representation for the image-text pair.

\noindent\textbf{Loss functions:} Two distinct loss functions were designed to train the proposed FLIP model: (1) Image-Text Contrastive loss (ITC): It aims to align the feature spaces of facial images and caption text \cite{DBLP:conf/icml/RadfordKHRGASAM21}. (2) Image-Text Matching loss (ITM): It is intended to facilitate the learning of image-text multimodal representations and align fine-grained features between image and text. ITM essentially represents a classification task, where the model predicts whether an image-text pair matches via an ITM head \cite{pmlr-v162-li22n,DBLP:conf/nips/LiSGJXH21}. 

\noindent\textbf{FLIP Settings:} The visual encoder of FLIP adopts a 12-layer ViT-B/16 \cite{dosovitskiy2020vit} with a token dimension of 768, and the input is a 224 $\times$ 224 image (about 223M parameters). A learnable [CLS] token was added before these 196 embeddings, resulting in 197 embeddings. The text encoder adopts a 12-layer BERT \cite{DBLP:conf/naacl/DevlinCLT19} model with an embedding dimension of 768. We fixed the number of input text tokens to 65 and truncated or padded the input if it did not match. Finally, we projected the image [CLS] and text [CLS] tokens into a 256-dimensional space, computed the ITC loss, inputted the 197 feature embeddings of ViT-B/16 into the BERT model, replaced the [ENC] token with the text feature embedding [CLS] token, and computed the ITM loss through cross-attention. 

\noindent\textbf{Training FLIP:} We implemented FLIP using the PyTorch framework and conducted the training on 8 $\times$ NVIDIA A100 GPUs. First, regarding the hyperparameter settings, we used a batch size of 8 $\times$ 80 for the 8 $\times$ NVIDIA A100 GPU. Similar to ALBEF \cite{DBLP:conf/nips/LiSGJXH21}, we set the queue size to 61440 to preserve the text and image features of multiple historical samples, which helped improve the robustness of the model. The image and text encoders were initialized using ViT-B \cite{dosovitskiy2020vit} and BERT-base \cite{DBLP:conf/naacl/DevlinCLT19}, respectively. We employed the AdamW \cite{DBLP:conf/iclr/LoshchilovH19} optimizer with a weight decay of 0.05. The learning rate was initialized to 1e-6, warmed to 3e-5 in 2,000 steps, and the cosine decayed to 1e-6 over 15 epochs. Notably, we used mixed precision to accelerate the training and save memory.

\subsection{Text-Image Retrieval}
Facial image-text retrieval tasks involve finding an image (text) based on a given text (image). To demonstrate the superior performance of FaceCaption-15M on this task, we conducted evaluations on CelebA-Caption and MM-CelebA \cite{Xia_2021_CVPR} datasets. For CelebA-Caption, we used 162,770 samples for training and 19,962 samples for testing. For the MM-CelebA \cite{Xia_2021_CVPR} dataset, we used 24,000 training samples and 6,000 testing samples. We adopted the Top-K retrieval index, where R@5 and R@10 represent the top-5 and top-10 accuracy, respectively.

\noindent\textbf{Baselines:} We selected three benchmark pre-trained models for comparison: (1) ALIGN, which was trained on a large-scale dataset containing 1 billion image-text pairs. (2) BLIP \cite{pmlr-v162-li22n}, which was trained on LAION-400M \cite{DBLP:journals/corr/abs-2111-02114}; (3) CLIP \cite{pmlr-v139-radford21a}, which was trained on a dataset with 400M image-text pairs from the Internet. To ensure fairness, all models used the Base-16 ViT module.

\noindent\textbf{Implementation details:} We froze the weights of the image encoders and text encoders for all pre-trained models, with initializing and fine-tuning two new linear layers as their output layers (as shwon in \Cref{fig7}(b)), the feature vectors of the output layers were used for the final retrieval. The training settings for these linear layers were consistent: employing the AdamW \cite{DBLP:conf/iclr/LoshchilovH19} optimizer, an initial learning rate of 4e-4, a weight decay of 0.05, and a learning rate schedule that followed cosine decay, to adjust the learning rate to 0 over 20 epochs.

\noindent\textbf{Performance Analysis:} From \Cref{table9}, we can observe: (1) The performance of our FLIP-based model, pretrained on FaceCaption-15M, achieves state-of-the-art results on both CelebA-Caption and MM-CelebA datasets. The performance is significantly improved compared to the models trained on other datasets, indicating the robust generalization ability of this dedicated pre-trained model. (2) The performance of our proposed is significantly better than that of FLIP$\dagger$, which verifies the lower text-image correlation in the LAION-Face dataset. (3) Our proposed also outperforms FLIP*, indicating the effectiveness of using the LLM model to rewrite raw text. These experiments further prove the necessity of constructing large-scale face text datasets.

\begin{table}[!t]
	\footnotesize
	\centering
	\caption{Comparison with other classical pretrained models. All pretrained model backbones are frozen, with only the linear layer being fine-tuned. $\dagger$ represents the model pretrained on the LAION-Face \cite{zheng2022general} dataset; * represents the model pretrained on the FaceCaption dataset constructed without using LLM text generation.}
	\setlength{\tabcolsep}{6pt}{
		\begin{tabular}{@{}c|cccc|cccc@{}}
			\toprule	
			& \multicolumn{4}{c|}{CelebA-Caption}                                                          & \multicolumn{4}{c}{MM-CelebA}                                                               \\
			Pretrained
			& \multicolumn{2}{c}{Text=\textgreater{}Image} & \multicolumn{2}{c|}{Image=\textgreater{}Text} & \multicolumn{2}{c}{Text=\textgreater{}Image} & \multicolumn{2}{c}{Image=\textgreater{}Text} \\
			Models& R@5                   & R@10                 & R@5                   & R@10                 & R@5                  & R@10                  & R@5                   & R@10                 \\ \midrule
			ALIGN \cite{DBLP:conf/nips/LiSGJXH21}      & 6.99                  & 11.09                & 5.08                  & 8.65                 & 8.1                  & 12.61                 & 6.6                   & 10.53                \\
			BLIP \cite{pmlr-v162-li22n}       & 24.56                  & 34.09                 & 24.35                  & 34.15                 & 21.56                 & 31.25                   & 23.83                  & 33.38                  \\
			CLIP \cite{pmlr-v139-radford21a}       & 21.42                 & 30.3                 & 26.52                 & 37.22                & 13.63                & 21.06                 & 14.41                 & 22.48                \\
			FLIP*            & {38.62}           & {48.28}          & {38.14}           & {48.9}           & {23.81}          & {33.48}           & {23.7}            & {32.83}          \\
			FLIP$\dagger$             & 13.24                 & 20.96                & 10.76                 & 17.08                & 9.75                 & 15.7                  & 7.48                  & 12.25                \\
			\textbf{FLIP (ours)}     & \textbf{50.04}        & \textbf{62.04}       & \textbf{50.96}        & \textbf{62.98}       & \textbf{26.1}        & \textbf{35.75}        & \textbf{27.05}        & \textbf{37.23}       \\ \bottomrule
		\end{tabular}%
	}

	\label{table9}
\end{table}

\subsection{Facial Attributes Prediction}
This task involves predicting attributes of a given facial image, such as gender, hairstyle, and so on, which can be regarded as a multilabel classification task. It has a wide range of applications in fields such as recommendation systems and security monitoring. To verify the effectiveness of our FLIP on this task, we conducted evaluations on the CelebA \cite{liu2015faceattributes} and LFWA \cite{5674057} datasets. For CelebA \cite{liu2015faceattributes}, we utilized 162,770 samples for training and 19962 samples for testing. For LFWA \cite{5674057}, we employed 6263 samples for training, reserving the rest for testing. Average precision (AP) was used as the evaluation index.

\noindent\textbf{Fine-tuning FLIP.} For this task, we only fine-tuned the final linear layer of FLIP. As shown in \Cref{fig7}(c), we frozen image encoder and used a randomly initialized new projection head as the output layer. This projection head utilized the features from $K$ layers of the ViT module, where K=\{4,6,8,12\}. We obtained three types of feature vectors: (1) the [CLS] token feature vector; (2) the average feature vector of all non-[CLS] tokens; (3) the max pooling feature vector of all non-[CLS] tokens. And then, we perform layer-normalize the 12 feature vectors and linearly combine them into one feature vector, followed by a fully-connected layer. Using AdamW \cite{DBLP:conf/iclr/LoshchilovH19} optimizer, an initial learning rate of 0.3, and a weight decay of 0.05, with cosine decay to 0 after 100 epochs.

\begin{table*}[!t]
	\setlength{\tabcolsep}{6pt}
	\caption{Comparison with other classical models. $\dagger$ represents the model pre-trained on the original LAION-Face dataset.}
	\footnotesize
	\centering
	\begin{tabular}{c|cccccc}
			\toprule[1.2pt]
			\multirow{2}{*}{}  & \multicolumn{3}{c}{\textbf{CelebA}}                  & \multicolumn{3}{c}{\textbf{LFWA}}               \\ \cline{2-7} 
			& \textbf{1\%}   & \textbf{2\%/10\%} & \textbf{100\%} & \textbf{10\%} & \textbf{50\%}  & \textbf{100\%} \\ \toprule[1.2pt]
			BEIT \cite{DBLP:conf/iclr/Bao0PW22}         & 85.64          & /88.74        & 89.71          & 71.06             & 70.99              & 70.98              \\
			ViT \cite{dosovitskiy2020vit}          & 89.20           & /90.21        & 90.77          & 79.58             & 85.19              & 84.16              \\
			CLIP \cite{pmlr-v139-radford21a}         & 89.09          & /90.47        & 90.86          & 80.20             & 85.48              & 84.48              \\ \hline
			SlimCNN \cite{DBLP:conf/fgr/SharmaF20}      & 80.96          & 82.32              & 91.24          & 71.49         & 73.45          & 76.02          \\
			DeepCluster \cite{DBLP:conf/eccv/CaronBJD18}  & 87.46          & 88.86              & 91.68          & 77.42         & 84.27          & 85.90           \\
			JigsawPuzzle \cite{DBLP:conf/eccv/NorooziF16} & 86.25          & 87.77              & 91.57          & 77.01         & 83.29          & 84.96          \\
			Rot \cite{DBLP:conf/iclr/GidarisSK18}          & 87.67          & 88.82              & 91.69          & 76.67         & 84.90           & 85.72          \\
			FixMath \cite{sohn2020fixmatch}      & 85.77          & 86.14              & 89.78          & 72.78         & 80.87          & 83.84          \\
			VAT \cite{miyato2018virtual}          & 86.30           & 87.28              & 91.44          & 74.42         & 80.55          & 84.68          \\
			SSPL \cite{shu2021learning}         & 88.84          & 89.58              & 91.77    & 81.65         & 85.43    & 86.53          \\
			FaRL \cite{zheng2022general}        & 89.66    & 90.26        & 91.39          & 82.42   & 85.38          & 85.94    \\ \hline
			FLIP$\dagger$-based     & 88.05 & /90.74     & 91.32 & 79.89 &85.62 & 84.34 \\ 
			\textbf{FLIP-based (ours)}      & \textbf{91.52} & \textbf{91.74}     & \textbf{92.27} & \textbf{82.60} & \textbf{89.87} & \textbf{98.04} \\ \toprule[1.2pt]
		\end{tabular}
	
	\label{Table2}
\end{table*}

\noindent\textbf{Performance Analysis.} In the case of the CelebA \cite{liu2015faceattributes} and LFWA \cite{5674057} datasets, we adopted the same ways with \cite{zheng2022general} that three different scale training subsets (1\%, 2\%/10\%, and 100\% for CelebA and 10\%, 50\%, and 100\% for LFWA) to fine-tune the linear layer of the FLIP model and subsequently evaluated its performance on the testing dataset. As shown in \Cref{Table2}, we can observe that: (1) The performance gradually improves with the increase in the amount of training data. (2) Interestingly, FLIP-based model exhibits a greater advantage when fine-tuned with less data. For example, the improvement of the FLIP model on LFWA is greater than that on CelebA when using a smaller percentage of data for fine-tuning. (3) Despite fine-tuning only the linear layer of FLIP, without further optimization of these features, our proposed model outperforms all the listed models trained on other datasets. These experiments demonstrate the capacity of FaceCaption-15M to enable the learning of more effective facial image representations. When combined with a well-crafted model, there is the potential for even greater improvement in this task.

\subsection{Sketch Less Facial Image Retrieval (SLFIR)}
Traditional sketch-based facial image retrieval (SBIR) depend on the high-quality sketch \cite{sain2022sketch3t,sain2023exploiting,chaudhuri2022bda}, which hinders its widespread applicability. SLFIR \cite{10095094} framework was designed to break these barriers by dynamic interactions during the sketching process. Its goal is to retrieve the image using a partial (poor) sketch with as few strokes as possible. For SLFIR, the main challenge is that the partial sketch often contains only localized information. In this section, we develop a multimodal learning method based on FLIP for the SLFIR problem.

\noindent\textbf{Dataset.} Based on FS2K dataset \cite{Fan2022FS2K}, Dai et al. \cite{10095094} constructed FS2K-SDE1 and FS2K-SDE2 datasets to address SLFIR problem. For FS2K-SDE1, 75,530 sketches and 1,079 images were used for training, the rest being used for testing. For FS2K-SDE2, 23,380 sketches and 334 images were used for training, the rest being used for testing. Owing to the significant differences in sketching among painters, Dai et al. \cite{10095094} constructed User-SDE that invited 50 professional painters to submit 110 sketch-drawing episodes. Following the method in \Cref{3}, we generated corresponding caption for the three datasets.

\noindent\textbf{Fine-tuning FLIP.} As shown in \Cref{fig7}(d), we frozen the image and text encoders, and initialized two linear layers as the output layers of the image and the text encoders, respectively. The fused feature vector obtained by adding the sketch and text feature vectors was used for the final retrieval. And we make the training settings of all linear layers consistent: using the AdamW \cite{DBLP:conf/iclr/LoshchilovH19} optimizer with an initial learning rate of 1e-06 and a weight decay of 0.05, the learning rate is adjusted to 0 by cosine decay within 100 epochs. The specific architecture of the fine-tuning model is illustrated in Figure D1. As mentioned in \cite{10095094}, \emph{m}@A (the ranking percentile) and \emph{m}@B (1/rank versus percentage of sketch) were used to capture the retrieval performance for the partial sketches.

\begin{table}[!t]
	\setlength{\tabcolsep}{6pt}
	\caption{Comparative results with different baseline methods. $\dagger$ represents the model pre-trained on the LAION-Face dataset. }
	\footnotesize
	\centering
	\begin{tabular}{c|cc|cc|cc}
		\toprule[1pt]
		& \multicolumn{2}{c|}{\textbf{FS2K-SDE1}} & \multicolumn{2}{c|}{\textbf{FS2K-SDE2}} & \multicolumn{2}{c}{\textbf{User-SDE}} \\ \toprule[1pt]
		Model         & \textbf{\emph{m}@A}        & \textbf{\emph{m}@B}      & \textbf{\emph{m}@A}        & \textbf{\emph{m}@B}      & \textbf{\emph{m}@A}       & \textbf{\emph{m}@B}      \\ \hline
		B1 \cite{DBLP:conf/cvpr/BhuniaYHXS20}    & 84.77               & 32.69             & 77.83               & 24.59             & 85.38              & 31.29             \\
		B2 \cite{DBLP:conf/cvpr/BhuniaYHXS20}             & 94.16               & 28.58             & 89.77               & 34.14             & -                  & -                 \\
		RL-B \cite{DBLP:conf/cvpr/BhuniaYHXS20}           & 84.42               & 22.76             & 85.65               & 26.70              & -                  & -                 \\
		SLFIR \cite{10095094} & 96.22               & 45.48             & 90.22               & 41.55             & 93.27              & 38.91             \\
		BLIP-based \cite{pmlr-v162-li22n}  & 97.64               & 46.72             & 97.10               & 64.06             & 94.68              & 48.99             \\
		CLIP-based \cite{pmlr-v139-radford21a}  & 98.01               & 57.27             & 96.02               & 57.32             & 95.57              & 58.38             \\ \hline
		FLIP$\dagger$-based & 98.14               & 56.70              & 95.41               & 59.40              & 89.69              & 38.89             \\
		\textbf{FLIP-based (ours)} & \textbf{99.49}      & \textbf{71.10}     & \textbf{97.55}      & \textbf{64.80}     & \textbf{97.34}     & \textbf{62.52}    \\ \toprule[1pt]
	\end{tabular}%
	
	\label{Table3}
\end{table}

\begin{figure*}[!t]
	\centering
	\includegraphics[width=1.0\linewidth]{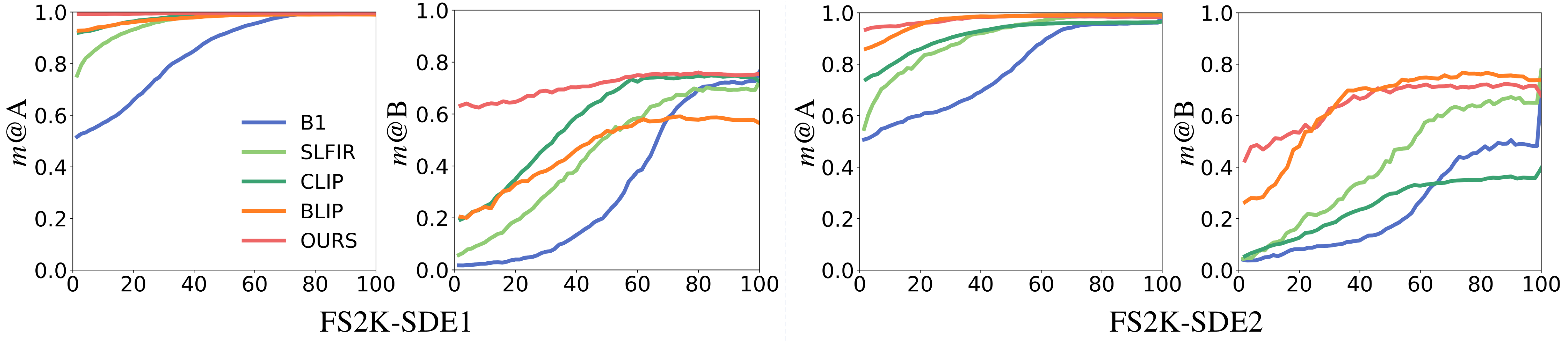}	
	\caption{Performance of early retrieval in SLFIR problem. Instead of showing the complete sketch, we visualized it using the percentage of sketch. A higher value indicates a better early retrieval performance.}
	\label{fig8}
\end{figure*}

\noindent\textbf{Performance Analysis.} The performance of our FLIP-based model on the SLFIR problem is shown in \Cref{fig8} and \Cref{Table3}. We can observe that: (1) Compared to single-modal sketch representations (including B1 \cite{DBLP:conf/cvpr/BhuniaYHXS20}, B2 \cite{DBLP:conf/cvpr/BhuniaYHXS20}, RL-B \cite{DBLP:conf/cvpr/BhuniaYHXS20} and SLFIR \cite{10095094}), the multimodal representation of facial sketches and text (including BLIP\cite{pmlr-v162-li22n}, CLIP \cite{pmlr-v139-radford21a}, FLIP$\dagger$, FLIP) can significantly improve the early retrieval performance on the SLFIR problem. (2) Our proposed FLIP model, trained on the FaceCaption-15M, achieves state-of-the-art results, and outperforms all methods at each stage of drawing process by a considerable margin. We believe that models carefully designed based on FaceCaption-15M have greater potential to further enhance performance on SLFIR.

\begin{figure}[!t]
	\centering
	\includegraphics[width=1\linewidth]{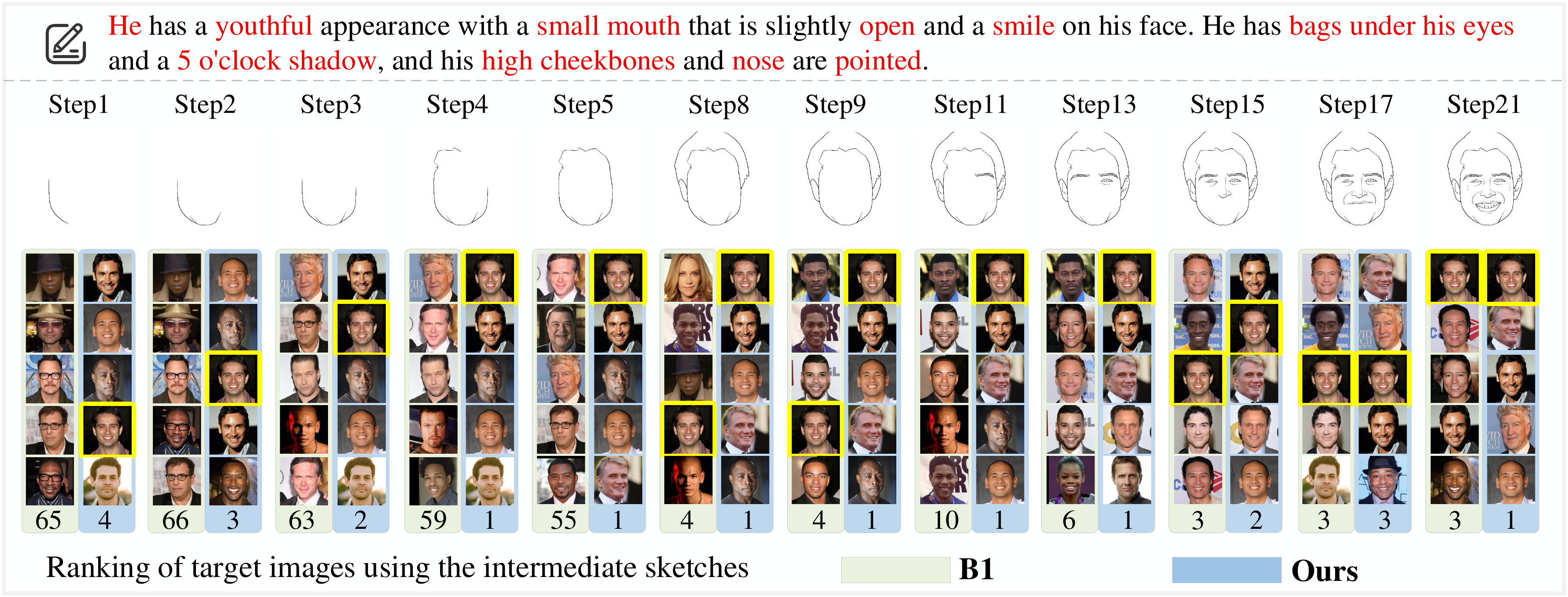}	
	\caption{Demonstration of our FLIP-based model on the SLFIR task. Both methods can retrieve the target face photo from the top-5 list using a partial sketch. Our proposed FLIP-based model can achieve this using fewer strokes than the baseline. The number at the bottom denotes the rank of the paired (true match) photos at every stage.}
	\label{fig9}
\end{figure}

\noindent\textbf{Practical Testing.} Considering the differences sketching among painters, we directly evaluated the models that were trained with FS2K-SDE1 on User-SDE. As shown in \Cref{Table3}, we can observe that: the performance of all models on User-SDE dataset decreases owing to the diversity of the sketching process, but our proposed FLIP-based model exhibits better generalization ability, and achieves state-of-the-art performance. Particularly, FLIP model has a significant advantage over the FLIP$\dagger$ model that trained on the LAION-Face \cite{zheng2022general} dataset, indicating the effectiveness of FaceCaption-15M. One illustration was shown in \Cref{fig9}, the target face image appears in top-5 list with only a very small amount of strokes incorporating with text information.

\section{Discussion}
We have proposed FaceCaption-15M, a comprehensive facial image-text dataset featuring 15 M image-text pairs with high-quality and diverse natural language descriptions of facial attributes. Our rigorous statistical analysis and experimentation demonstrated the superior and quality effectiveness of FaceCaption-15M. In future, we have plans to further improve both the image quality and richness of text within FaceCaption-15M. We also plan to explore new tasks and develop new models based on this dataset. We will make the dataset available after undergoing a rigorous legal check at our institution.

\noindent\textbf{Ethical Consideration:} We only annotated generic attributes such as gender, hair color, and wears. Additionally, facial images and captions generated in this work are devoid of bias or certain biometric information, alleviating ethical concerns. We are committed to carefully controlling the application and acquisition procedures for FaceCaption-15M to prevent any potential misuse or abuse.

We are aware that the scale of FaceCaption-15M could have dual impacts on society, both advancing facial recognition technology and digital content innovation, and potentially being misused, such as for Deepfake fraud.
In terms of fairness, thanks to the comprehensiveness of the LAION dataset and our strict requirements for fairness during the dataset creation process, FaceCaption-15M demonstrates a high level of fairness. We also recognize the limitations in achieving absolute fairness with the current dataset, and the selection of displayed images may exacerbate this impression of unfairness. Additionally, in the subsequent maintenance of the dataset, we will continuously introduce new technologies to improve data annotation and reduce linguistic biases in the dataset.

%
%
\bibliographystyle{splncs04}
\bibliography{main}
\end{document}